\documentclass[conference]{IEEEtran}
\usepackage{times}
\usepackage{latexsym}
\usepackage{amsmath,amssymb}
\usepackage{bm}
\usepackage[T1]{fontenc}
\usepackage[utf8]{inputenc}
\usepackage{algorithm}
\usepackage{algorithmic}
\usepackage{multirow}
\usepackage{graphicx}
\usepackage{adjustbox}
\usepackage{microtype}

\DeclareMathOperator*{\argmax}{argmax}

\def\BibTeX{{\rm B\kern-.05em{\sc i\kern-.025em b}\kern-.08em
    T\kern-.1667em\lower.7ex\hbox{E}\kern-.125emX}}
\begin{document}

\title{Federated Non-negative Matrix Factorization for Short Texts Topic Modeling with Mutual Information}
\author{\IEEEauthorblockA{Shijing Si$^{1}$, Jianzong Wang$^{1*}$, Ruiyi Zhang$^{2}$, Qinliang Su$^{3}$ and Jing Xiao$^{1}$}
  \IEEEauthorblockN{$^{1}$Ping An Technology (Shenzhen) Co., Ltd., Shenzhen, China\\
  $^{2}$Duke University, Durham, USA \\
  $^{3}$Sun Yat-sen University, Guangzhou, China}
\IEEEauthorblockN{\thanks{* Corresponding author: Jianzong Wang, \texttt{jzwang@188.com}.}}}

\maketitle

\begin{abstract}
Non-negative matrix factorization (NMF) based topic modeling is widely used in natural language processing (NLP) to uncover hidden topics of short text documents.
Usually, training a high-quality topic model requires large amount of textual data. In many real-world scenarios, customer textual data should be private and sensitive, precluding uploading to data centers. This paper proposes a Federated NMF (FedNMF) framework,
which allows multiple clients to collaboratively train a high-quality NMF based topic model with locally stored data. 
However, standard federated learning will significantly undermine the performance of topic models in downstream tasks (\textit{e.g.}, text classification) when the data distribution over clients is heterogeneous. To alleviate this issue, we further propose FedNMF+MI, which simultaneously maximizes the mutual information (MI) between the count features of local texts and their topic weight vectors to mitigate the performance degradation. 
Experimental results show that our FedNMF+MI methods outperform Federated Latent Dirichlet Allocation (FedLDA) and the FedNMF without MI methods for short texts by a significant margin on both coherence score and classification F1 score. 
\end{abstract}


\section{Introduction}
Topic models \cite{steyvers2007probabilistic,alghamdi2015survey} provide exploratory approaches to analyze large volumes of unlabeled text, which could discover clusters of tokens that co-occur often and organize large
capacity of documents. They have many applications in text mining and analysis \cite{wosik2019artificial,al2020mining}.  
In practice, most commonly used topic models can be grouped into two types: generative methods like Latent Dirichlet
Allocation (LDA) \cite{blei2003latent,xue2020public} and Non-negative Matrix Factorization (NMF) \cite{paatero1994positive,luo2017probabilistic,li2021topic}.
A few studies \cite{suri2017comparison,chen2019experimental} have proven that NMF outperforms LDA for short texts and is faster than the latter. Also, there are completely deterministic algorithms for its resolution \cite{lee2009semi,kim2014algorithms}. Therefore, NMF and its variants are commonly used to extract topics from short-text datasets \cite{shi2018short,shahbazi2020topic}.
Topic models rely on large amount of textual data. Meanwhile, since data could be proprietary and sensitive, regulations such as the recently enforced European Union General Data Protection Regulation (GDPR) \cite{carey2018data} may preclude uploading them to 
data centers and being utilized in the conventional centralized approach. Federated learning (FL) has been proposed to collaboratively train a machine learning model over decentralized data on clients \cite{mcmahan2017communication,xu2021federated}.
To address this issue for topic modeling, a few research investigate how to implement LDA in a federated manner \cite{jiang2019federated,shi2020federated1}.


However, how to implement NMF based topic modeling in a federated manner is still underexplored. To fill this gap, we present a framework, called federated NMF (FedNMF), which enables multiple clients to jointly train a NMF based topic model.
As shown in Table \ref{tab:clf} of our experiments (Section \ref{sec:exp}), 
simply implementing FedNMF with existing FL algorithms like FedAvg and FedAdam \cite{mcmahan2017communication,reddi2021adaptive}, the performance of downstream text classification may degrade significantly for heterogeneous data over clients, in comparison with the centralized setting. To address this problem, we propose
to maximize the mutual information \cite{belghazi18mutual,zhelezniak2020estimating} between the original bag-of-word representations of documents and their topic weight vectors, which can be treated as high-level representations of the raw documents.
 We introduce mutual information to topic modeling and show its effectiveness through experiments. The reasoning behind this technique is that topic weight vectors of the right factor matrix can be treated as a high-level representation of each document.


The contributions of this work are summarized as follows:
\begin{itemize}
    \item We propose a federated framework, FedNMF, which enables multiple clients to train a NMF based topic modeling with stochastic gradient descent (SGD).
    \item To further boost the performance of FedNMF, we propose FedNMF+MI, which maximizes the mutual information between text features and its high-level representations.
    \item We evaluate our framework on a few public short-text datasets to verify its effectiveness. Also we investigate its performance under varying federated settings.
\end{itemize}

\section{Related Work}
Our work is related to three lines of research, which
are federated learning, NMF based topic modeling and mutual information.

\paragraph{Federated Topic Modeling}

FL, proposed by \cite{mcmahan2017communication}, aims to efficiently train a model by leveraging decentralized data over multiple clients.
Since its inception, many algorithms such as FedAvg \cite{mcmahan2017communication} and FedOpt \cite{reddi2021adaptive} have been developed to reduce communication cost while retaining the performance. These algorithms usually proceed in two steps: (1.) synchronize the latest model weights from the server and train the model with local data; and (2.) after training, each client transmits the updated weights to the server for aggregation. These advanced FL algorithms are compatible to our FedNMF framework.

A few research has devoted to federated topic modeling.
For example, \cite{jiang2019federated} presented federated topic modeling that
consists of novel techniques such as private Metropolis
Hastings, topic-wise normalization and heterogeneous model integration. 
\cite{shi2020federated1} developed federated LDA that is based on a novel local differential privacy (LDP)
\cite{evfimievski2003limiting}
mechanism. These works mainly focus on the LDA based topic modeling. By contrast, in this paper we study the NMF based topic modeling under federated settings.

\paragraph{Mutual information}

The mutual information between two random variables $X$ and $Y$ is the KL divergence between the
joint and the product of marginals.
$$I(X; Y) = \mathcal{D}_{KL}(P(X, Y)||P(X)P(Y)),$$
which we hope to estimate with samples from the joint probability $P(X, Y)$. 
Mutual information measures the 
inherent correlation between random variables (vectors).
It has been widely used in unsupervised deep learning to improve representation learning for images, texts, etc \cite{belghazi18mutual,poole2019variational,Kong2020A,cheng2020club,cheng2021fairfil}. Here we utilize the state-of-the-art mutual information estimator to enhance our FedNMF framework.

\section{Methodology}

Here we introduce our methodology. We use regular uppercase letters to denote matrices and boldface lowercase letters to denote vectors. 
For example, $\mathbf{A}$ is a $m$-by-$n$ non-negative real matrix, whose element $(i,j)$ is denoted by $\mathbf{A}_{i,j}$. 
More details on mathematical notations can be found in Table \ref{tab:notation}.

For a matrix $\mathbf{A}$, NMF aims to find two find two non-negative factor matrices $\mathbf{W}$
and $\mathbf{H}$ such that $\mathbf{A}\approx\mathbf{W}\mathbf{H}$. One straightforward way is to minimize the following $\ell_2$ loss
with respect to $\mathbf{W}$
and $\mathbf{H}$.
{
\begin{align}
    {L}(\mathbf{A}; \mathbf{W}, \mathbf{H})=&\sum_{(i,j)}(\mathbf{A}_{i,j} - \hat{\mathbf{A}}_{i,j})^2 \nonumber\\=&||\mathbf{A} - \mathbf{WH}||_2^2.\label{eq:square_loss}
\end{align}
}%

\begin{table*}[htp]
\centering
\caption{Table of Mathematical Notations}
\vspace{-3mm}
\begin{adjustbox}{scale=1.,tabular=l p{6.1cm} |l p{5.1cm},center}\\
\hline
\textbf{Notation}  & \textbf{Property} & \textbf{Notation}  & \textbf{Property}  \\
\hline
$k$   & The number of topics \emph{i.e.}, the rank of matrix $\mathbf{W}$ &
$V$  & The size of vocabulary     \\
$K$  & Total number of clients     &
$C$  & Fraction of clients used for each iteration    \\
$N_{i}$  & The number of documents on the $i$-th Client  &
$N=\sum_{i}N_{i}$  & Total number of documents on all clients \\
$\mathbf{A}=(\mathbf{A}_{1}, \ldots, \mathbf{A}_{K})$       & Whole token-document matrix    &
$\mathbf{A}_{i,j}$       & The $(i,j)$-th entry of matrix $\mathbf{A}$    \\
$\mathbf{A}_{i}$       & Token-document matrix of the $i$-th client  &
$\mathbf{A}_{i}(:,j)$       & The $j$-th column of matrix $\mathbf{A}_{i}$  \\
$\mathbf{H}_{i}$       & Topic-document matrix of the $i$-th client  &
$\mathbf{H}_{i}(:,j)$       & The $j$-th column of matrix $\mathbf{H}_{i}$  \\
$\mathbf{W}$       & Token-topic matrix    &
$\mathbf{H}=(\mathbf{H}_{1}, \ldots, \mathbf{H}_{K})$    & Whole topic-document matrix     \\
$T_{\theta}(\cdot, \cdot)$  & The neural networks used in
 mutual information &
$E$  & Epochs of SGD on local clients \\
& estimator with parameter $\theta$ & $B$  & Mini-batch size on local clients \\
\hline
\end{adjustbox}
\vspace{-5mm}
\label{tab:notation}
\end{table*}

\begin{figure}[h]
\centering
\includegraphics[width=\linewidth]{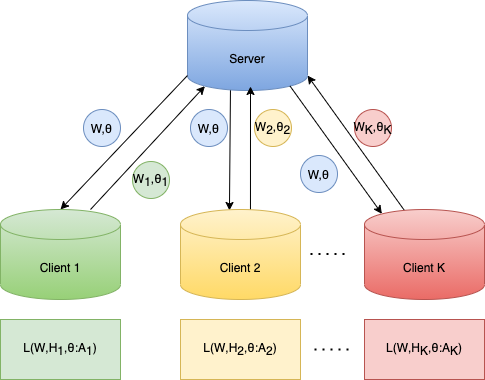}
\caption{Federated learning framework for NMF. Data are locally stored over client devices, and model weights are updated on each client, uploading to the central server for aggregation.}
\label{fig:fed_nmf}
\end{figure}
For each document, it can be represented by a count (column) vector on top of bag-of-word representation. For documents on
the $i$-th client, we denote the count feature matrix as $\mathbf{A}_{i}$. The concatenation of $(\mathbf{A}_{i}, i= 1, \ldots, K)$ is the matrix to be decomposed.

\subsection{Federated NMF}

Figure \ref{fig:fed_nmf}
depicts how FedNMF factorizes multiple matrices distributed over different clients.
In the federated learning framework, we use the stochastic gradient descent (SGD) algorithm to minimize the loss in Eq. \eqref{eq:square_loss}.
However, merely optimizing the squared loss on each client inside the FedAvg algorithm
will lead to poor topic models, which is verified in the experiments in Section \ref{sec:exp.result}.
Our intuition is to maximize the mutual information
between the original textual features $\mathbf{A}_i$ and features in the factor matrix $\mathbf{H}_i$. Directly estimate the mutual information is not tractable, and we bypass this difficulty utilizing the SMILE lower bound of mutual information \cite{Song2020Understanding}  
defined as
\begin{multline*}
  I_{\rm SMILE}(T_{\theta}, \tau)=\sup\limits_{T_{\theta}}
    \mathbb{E}_{P(\mathbf{x}, \mathbf{y}))}[T_{\theta}(\mathbf{x}, \mathbf{y})] -\\ \log\mathbb{E}_{Q}[\text{clip}(e^{T_{\theta}(\mathbf{x}, \mathbf{y})}, e^{-\tau}, e^{\tau})]\,,  
\end{multline*}
where $\text{clip}(v, l, u) = \max(\min(v, l), u)$, $\tau > 0$ is a hyper-parameter (fixed at $5.0$ in this paper) and ${P}, {Q}$ are the joint and marginal distributions of $(\mathbf{x}, \mathbf{y})$, respectively.

On a mini-batch (denoted as $S$, a set of column indices) of text documents on client $i$ with batch size $B$, the SMILE estimator is 
\begin{equation}\label{eq:smile}
\begin{split}
&\hat{I}_{\rm SMILE}(\theta,\mathbf{H}_{i}, \mathbf{A}_i,S) = \frac{1}{B}\sum_{j\in{S}}T_{\theta}(\mathbf{A}_{i}(:,j), \\ &\mathbf{H}_{i}(:,j))-\log\Big(\frac{1}{B(B-1)}
\sum\limits_{j,j'\in{S}, j\neq j'}\text{clip}\Big(\\
&e^{T_{\theta}(\mathbf{A}_{i}(:,j), \mathbf{H}_{i}(:,j'))}, e^{-\tau}, e^{\tau}\Big).
\end{split}
\end{equation}
Maximizing the mutual information estimator in Eq. \eqref{eq:smile} forces the high-level features in the matrix $\mathbf{H}_i$ to capture information in the original
term-document matrix $\mathbf{A}_i$.

Plus the least square loss, the total loss on a certain mini-batch $S$ of client $i$ is
\begin{align}\label{eq:total.loss}
L(\mathbf{W},\mathbf{H}_{i}, \theta; \mathbf{A}_{i}, S)
&=-
\lambda\cdot\hat{I}_{\rm SMILE}(\mathbf{W},\mathbf{H}_{i}; \mathbf{A}_i,S) \nonumber \\+\frac{1}{B}\sum_{j\in S}||\mathbf{A}_{i}&(:,j) - \mathbf{W}\mathbf{H}_{i}(:,j)||_{2}^{2},
\end{align}
in which $\lambda\geq 0.0$ controls the strength of MI regularizer.
When training the model on the $i$-th client, the parameter $\theta$ of mutual information estimator and matrix decomposition are separately learned, \emph{i.e.},
\begin{align*}
    &\hat{\theta}=\argmax_{\theta}\hat{I}_{\rm SMILE}(\mathbf{W},\mathbf{H}_{i}; \mathbf{A}_i,S)\\
    &(\hat{\mathbf{W}}, \hat{\mathbf{H}_i})=\argmax_{\mathbf{W}, \mathbf{H}_i}L(\mathbf{W},\mathbf{H}_{i}, \hat{\theta}; \mathbf{A}_{i}, S)
\end{align*}

The complete pseudo-code of our FedNMF+MI framework is shown in Algorithm \ref{alg:fed.nmf}. 
When the MI regularizer weight $\lambda = 0.0$, this algorithm degenerates to FedNMF without MI, or plainly FedNMF.

\begin{algorithm}[tb]
\caption{FedNMF+MI. The $K$ clients are indexed by $i$; $B$ is the local mini-batch size, $E$ is the number of local epochs, and $\eta$ is the learning rate.}
\label{alg:fed.nmf}
\begin{algorithmic}[1] 
\STATE \textbf{Server Executes}: 
\STATE \textbf{Initialize} $\mathbf{W}^{(0)}$ and $\theta^{(0)}$
\FOR{each round $t = 1, 2, \ldots$}
\STATE $m\leftarrow\max(C\cdot K, 1)$
\STATE $G_t\leftarrow$(random group of $m$ clients)
\FOR{each client $i\in G_t$ \textbf{in parallel}}
\STATE $(\mathbf{W}_{i}^{(t+1)}, \theta_{i}^{(t+1)})\leftarrow\text{ClientUpdate}(i, \mathbf{W}^{(t)}, \theta^{(t)})$
\ENDFOR
\STATE $\mathbf{W}^{(t+1)}\leftarrow\textbf{Agg}(\{\mathbf{W}_{i}^{(t+1)}, i\in{G}_t\})$
\STATE ${\theta}^{(t+1)}\leftarrow\textbf{Agg}(\{{\theta}_{i}^{(t+1)}, i\in{G}_t\})$
\STATE \textbf{Agg} is the aggregation rule, it can be weighted average (FedAvg) or adaptive optimization (FedOpt)
\ENDFOR
\end{algorithmic}
\vspace{2mm}
\begin{algorithmic}[1]
\STATE \textbf{ClientUpdate$(i, \mathbf{W}, \theta)$}:~~//Run on client $i$
\FOR{each local epoch from $1$ to $E$}
\STATE $\mathcal{B}\leftarrow($ split column indices of $\mathbf{A}_i$ into batches of size $B$)
\FOR{batch $b\in\mathcal{B}$}
\STATE $\mathbf{W} \leftarrow\mathbf{W} - \eta\bigtriangledown{L}(\mathbf{W}, \mathbf{H}_{i}, \theta; \mathbf{A}_{i}, b)$, in which the loss $L$ is defined in Eq. \eqref{eq:total.loss}
\STATE $\mathbf{H}_{i} \leftarrow\mathbf{H}_{i} - \eta\bigtriangledown{L}(\mathbf{W}, \mathbf{H}_{i}, \theta; \mathbf{A}_{i}, b)$
\STATE $\theta \leftarrow\theta + \eta\bigtriangledown\hat{I}_{\rm SMILE}(\theta, \mathbf{H}_{i}; \mathbf{A}_{i}, b)$, \emph{i.e.}, optimizing $\theta$ by maximizing SMILE mutual information in Eq. \eqref{eq:smile}
\ENDFOR
\ENDFOR
\STATE return $(\mathbf{W}, \theta)$ to server
\end{algorithmic}
\end{algorithm}

In Algorithm \ref{alg:fed.nmf}, the \textbf{Agg}$(\cdot)$ is the aggregation function and it has many choices. The most basic one is weighted average that proceeds as follows
\begin{align*}
  &\mathbf{W}^{(t+1)}\leftarrow\sum_{i=1}^{K}\frac{N_{i}}{N}\mathbf{W}_{i}^{(t+1)} \\
&\theta^{(t+1)}\leftarrow\sum_{i=1}^{K}\frac{N_{i}}{N}\theta_{i}^{(t+1)}.
\end{align*}
This is the FedAvg algorithm, and our framework is also compatible to other aggregation algorithms like FedAdam, FedAdagrad and FedYogi. More details on these aggregation algorithms can be found in \cite{reddi2021adaptive}.

\section{Experiments}\label{sec:exp}

In this section we implement our algorithms over a few public datasets and we compare the performance of our methods with the state-of-the-art FedLDA to show the effectiveness of our methods. All the models are implemented using PyTorch \cite{paszke2019pytorch} and trained on
NVIDIA Tesla V100 GPUs. 

\subsection{Datasets}
The four real-world text
datasets in the experiments correspond to various types of applications, \textit{i.e.}, Emails, News,
Questions \& Answers, Microblogs.
\paragraph{Spam Ham} This data is a public set of labeled messages that have been collected for mobile phone spam research. It consists of 5,572 English, real and non-enconded messages, tagged according to being legitimate (ham) or spam.

\paragraph{Tag.News} This data set is a part of the Tag News dataset3,
which is composed of news, snippets and tweets \cite{zubiaga2013harnessing}.
The articles in the dataset belong to one of the following 7 categories: Business, Entertainment, Health, Sci\&Tech, Sport, US and World.
\paragraph{Rt Polarity} This dataset contains 5331 snippets for both positive and negative movie reviews, respectively. It was published and used in \cite{pang2005seeing}.

\paragraph{Stack Overflow} It is a collection of $20,000$ questions posted on StackOverflow.com, and each question is associated to tags like ``html", ``R", ``python", etc. More details on this dataset can be found in \cite{baltes2020annotated}.



\begin{table}[t]
\caption{Basic statistics of datasets in the experiments \label{tab:stats}}
\vspace{-5mm}
\begin{adjustbox}{width=\linewidth,tabular=lllll,center}\\
\hline
Data Set  & \#docs & \#terms & len. & \#cats \\ \hline
Spam Ham & 5572   & 7076    & 8.56      & 2       \\
Tag.News  & 1434   & 12393   & 30.29      & 7      \\
Rt Polarity & 10662  & 16241  & 10.66      & 2     \\
Stack Overflow    & 20000  & 9316  & 5.44     & 20       \\\hline
\vspace{-4mm}
\end{adjustbox}
\end{table}
\begin{table}[t]
\caption{Average topic coherence on four datasets with
five topic settings [ 50, 100, 200] in terms of word embeddings (WE) based coherence score. The bold numbers represent the best performance in each category. \label{tab:coherence}}
\vspace{-5mm}
\begin{adjustbox}{width=\linewidth,tabular=lllll,center}\\
\hline
\multirow{2}{*}{Methods/Data}  & Spam & Tag. & Rt & Stack  \\
  & Ham & News & Polar. & Overf. \\
  \hline
LDA  & 0.404  & 0.386   & 0.404      & 0.302  \\
 NMF & \textbf{0.442}  & \textbf{0.486}   & \textbf{0.466}      & \textbf{0.324}   \\\hline
  NMF SGD & 0.421  &  0.482  &   0.453    &  0.326  \\
   NMF SGD+MI & \textbf{0.456}  &  \textbf{0.486}  &    \textbf{0.490}   & \textbf{0.340}  \\\hline
FedLDA      &   0.341     &   0.370      &            0.406&  0.306   \\
FedAvg    &   0.347     &     0.423    &        0.412    & 0.316    \\
FedAvg+MI    &   0.362     &     0.446    &  0.431  & 0.328    \\
FedAdagrad    &   0.347     &     0.423    &        0.411    & 0.313    \\
FedAdag.+MI    &   0.352     &     0.426   &  0.432  & 0.332    \\
FedYogi    &   0.349     &     0.423    &        0.416    & 0.317    \\
FedYogi+MI    &   0.355     &     0.446    &  \textbf{0.456}  & 0.332    \\
FedAdam    &   0.348     &     0.423    &        0.424    & 0.316    \\
FedAdam+MI    &   \textbf{0.372}     &     \textbf{0.466}    &  0.453  & \textbf{0.342}    \\
\hline
\vspace{-5mm}
\end{adjustbox}
\end{table}

\begin{table*}[!h]
\centering
\caption{Average F1 score and accuracy on four datasets with
three topic settings [50, 100, 200] of text classification task on top of topic modeling. The highlighted numbers represent the best performance in each category.}
\label{tab:clf}
\vspace{-5mm}
\begin{adjustbox}{width=0.9\linewidth,tabular=llrrlrrllllrr,center}\\
\hline
Dataset    &  & \multicolumn{2}{c}{Spam Ham}                          &           & \multicolumn{2}{c}{Tag.News}                                         &           & \multicolumn{2}{c}{Rt polarity}   &     & \multicolumn{2}{c}{Stack Overflow}          \\ \cline{3-4} \cline{6-7} \cline{9-10} \cline{12-13} 
Metrics    &  & \multicolumn{1}{l}{F1} & \multicolumn{1}{l}{Acc.} &           & \multicolumn{1}{l}{F1}                & \multicolumn{1}{l}{Acc.} &           & F1                                 & Acc.                           &           & \multicolumn{1}{l}{F1} & \multicolumn{1}{l}{Acc.} \\ \hline
LDA        &  & 0.821   & 0.903         &           & 0.651                  & 0.653         &           &   0.594    &      0.594       &           & 0.689   & 0.683       \\
NMF        &  & \textbf{0.904}                 & \textbf{0.954}                        &           &  \textbf{0.859}          & \textbf{0.840}                         &           & \textbf{0.605}  & \textbf{0.608}                            &           & \textbf{0.803}                  & \textbf{0.747}                       \\ \hline
NMF+SGD    &  & 0.927                  & 0.965                        &           & 0.950          & 0.969                        &           & 0.547          & 0.548          &           & 0.777                  & 0.778                      \\
NMF+SGD+MI &  & \textbf{0.941}         & \textbf{0.968}               &  &  \textbf{0.982} & \textbf{0.983}               & \textbf{} & \textbf{0.670} & \textbf{0.700} &  & \textbf{0.820}         & \textbf{0.831}                          \\ \hline
FedLDA     &  & 0.704   & 0.841         &           &  0.508     & 0.555         &           &   0.500    &    0.501   &           & 0.421   & 0.401      \\
FedAvg     &  & 0.652                  & 0.673                        &           & 0.663                                 & 0.670 &           & 0.540          & 0.541          &           & 0.304                  & 0.313                       \\
FedAvg+MI  &  & 0.724         & 0.730               &           & 0.724                        & 0.740       &           & 0.594 & 0.597 &  & 0.360         & 0.421   \\
FedAdagrad     &  & 0.816                  & 0.830                        &           & 0.920                                 & 0.931 &           & 0.546          & 0.550          &           & 0.558                  & 0.600                      \\
FedAdagrad+MI  &  & 0.856         & 0.867               &           & 0.953                        & 0.956               &           & 0.584 & 0.592 & & 0.630         & 0.641   \\
FedYogi     &  & 0.827                  & 0.837                        &           & 0.924                                 & 0.930 &           & 0.567          & 0.571          &           & 0.352                  & 0.360                       \\
FedYogi+MI  &  & 0.861         & 0.870               &           & \textbf{0.973}                        & \textbf{0.970}               &           & 0.598 & 0.605 &  & 0.372         & 0.367   \\
FedAdam     &  & 0.833                  & 0.838                        &           & 0.890                                 & 0.638 &           & 0.572         & 0.581          &           & 0.562                  & 0.584                       \\
FedAdam+MI  &  & \textbf{0.881}         & \textbf{0.887}               &           & 0.945                        & 0.957  &           & \textbf{0.618} & \textbf{0.622} &  & \textbf{0.647}         & \textbf{0.650}   \\\hline
\end{adjustbox}
\vspace{-5mm}
\end{table*}

Table \ref{tab:stats} illustrates the basic statistics of datasets. In this table, `\#docs', `\#terms' and `\#cats' represents the number of documents, terms and categories in each dataset, respectively. `doc-length' is the average length of documents in each dataset. All these figures are computed after stop-words removal. In our experiments, we split each dataset into training and test sets, taking 80\% and 20\%, respectively.

\subsection{Evaluation Metrics}
Following \cite{fang2016using} and \cite{newman2010automatic},
 we utilize the word embedding (WE) topic coherence to evaluate the quality of topics.
More importantly, we employ an external document classification \cite{si2020students,wang2020integrating} to measure the performance of topic modeling methods.

\paragraph{\textbf{WE Topic Coherence}} is defined as follows:
    \begin{equation}\label{eq:coherence}
        \mathcal{C} = \frac{2}{\mathcal{N}(\mathcal{N}-1)}\sum_{1\leq~i<j\leq\mathcal{N}}{Cosine}(w_i, w_j)
    \end{equation}
    where $(w_1, \ldots, w_{\mathcal{N}})$ is the top $\mathcal{N}$ number of most probable words for each topic. $Cosine(w_i, w_j)$ is the cosine similarity of the word embeddings for words $w_i$ and $w_j$, and here we use the 200-dimensional GloVe embeddings \cite{pennington2014glove} trained on the Wikipedia 2014 and Gigaword 5 corpora. The average WE coherence score of all topics measures the quality of each topic model.
\paragraph{\textbf{Document Classification}}
Another popular way to evaluate
the goodness of a topic model is to leverage the latent document representations for external tasks. In our experiments, we conduct text classification using the topic weights, i.e., the columns of $\mathbf{H}$ as features of documents.

\subsection{Comparison Methods}
Here are the list of topic modeling methods we implemented in our experiments from centralized methods to federated ones.

\begin{itemize}
    \item\textbf{Centralized Traditional Methods:} LDA and NMF are conventional topic models for centralized textual data.
    \item\textbf{Centralized SGD-based Methods:} We include SGD-based NMF methods (NMF+SGD and NMF+SGD+MI) in our experiments to verify the effectiveness of mutual information for the centralized NMF topic modeling. Essentially, for NMF+SGD, we optimize the least square loss of NMF with mini-batch SGD. An additional mutual information regularizer is included for NMF+SGD+MI.

    \item\textbf{Federated Topic Modeling:} We implement three federated topic modeling approaches: FedLDA based on variational inference \cite{jiang2019federated,shi2020federated1}, FedNMF with four implementation methods (FedAvg, FedAdagrad, FedYogi and FedAdam). We vary the value of $\lambda$ to include or exclude the mutual information term, i.e., setting $\lambda=0$ for plain FedNMF, and $\lambda > 0$ for FedNMF+MI. 
\end{itemize}
    

\subsection{Experimental Setup}

In this experiment, 
we construct datasets for various clients in the same way as \cite{hsu2019measuring}.
Specifically, we assume on every client training examples are drawn independently
with class labels following a categorical distribution over $l$ classes parameterized by a vector $\mathbf{q}$ with elements ($q_i\geq 0, i\in[1, l]$ and $\sum{q}_i = 1$). We synthesize a population of non-identical clients by drawing $\mathbf{q}\sim\text{Dir}(\alpha\mathbf{p})$ from a Dirichlet distribution, where $\mathbf{p}$ is the label distribution of a certain dataset, and $\alpha$ controls the identicalness among clients. With $\alpha\rightarrow{\infty}$, all clients have identical distributions to $\mathbf{p}$; with $\alpha\rightarrow{0}$, on the other extreme, each client holds examples from only one label. 

In our experiments, we use $\alpha$ to control the data heterogeneity among clients and vary client number $K$ to create different FL settings. 
Throughout the whole experiments, we set $K$ to the set \{1, 10, 30, 50\}, and to make our results comparable, we make each client to have equal number of documents, which is the total sample size divided by client number $K$.
After the federated topic modeling, we aggregate the topic weight vectors of all documents, randomly split it into train set (80\%) and test set (20\%), and train a support vector machine (SVM) classifier on top of them to yield the macro F1 score and accuracy.
We set the number of topics $k$ to four values 20, 50, 100 and 200 for all datasets to obtain comprehensive results.
For the federated topic modeling approaches, we set the fraction of participants for each round, $C=0.2$. At each iteration, we tune the number of epochs for local SGD training from the set \{10, 20, 30, 40\}, and local batch size from
the set \{16, 32, 64, 128\}, the learning rate from the set $\{1.e-2, 2.e-2, 5.e-2, 1.e-1\}$, the weight $\lambda$ from the set $\{0, 0.01, 0.05, 0.1, 0.5\}$.
The hyper-parameters in FedAdagrad, FedYogi and FedAdam are set as default in \cite{reddi2021adaptive}.
The neural networks $T_{\theta}$ used for mutual information is a 2-layered feed-forward networks with 32 and 256 neurons in the 2 hidden layers, respectively. 

\subsection{Results and Analysis}\label{sec:exp.result}

\begin{figure}[h]
\centering
\includegraphics[width=\linewidth, height=1.1\linewidth]{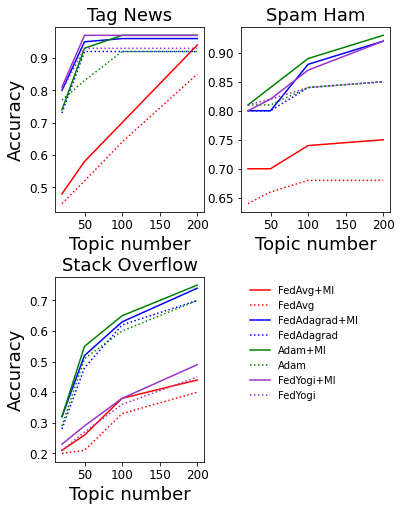}
\caption{The performance of various federated NMF topic modeling algorithms in different colors versus the topic number $k$ on three datasets: Tag.News, Spam Ham, and Rt polarity. FL setting: $\alpha=1$ and $K=10$. 
}
\label{fig:performance_topic_k}
\end{figure}

Table \ref{tab:coherence} and \ref{tab:clf} illustrate the coherence score and text classification metrics of various topic modeling approaches over four datasets. The reported coherence scores, F1 scores and accuracy values are averaged over three topic numbers $k=50, 100$ and $200$. The FL environment is set as $\alpha=1.0$ and $K=10$.
Here are the main observations: 1.) By comparing traditional LDA and NMF, we find that NMF outperforms LDA on all datasets, and these documents are short in general, smaller than 30 by length. Therefore, NMF usually outperforms LDA on short documents. 2.) Mutual information benefits NMF topic modeling under both centralized and federated settings. This is shown in both coherence score and classification. For instance, in Table \ref{tab:clf}, NMF+SGD+MI consistantly yields better F1 scores than NMF+SGD in centralized learning over four datasets, especially on Rt polarity, which presents a 15\% margin. This exhibits the effectiveness of mutual information in NMF based topic modeling. 3.) Among the federated topic models, FedNMF+MI methods (including FedAvg+MI, FedAdagrad+MI, FedYogi+MI, and FedAdam+MI) produce better results than FedLDA on all four datasets, though FedNMF without MI (FedAvg, FedAdagrad, FedYogi and FedAdam) perform slightly better than FedLDA. 

\paragraph{Classification over topic number} The text classification is closely related to the number of topics and Figure \ref{fig:performance_topic_k} presents the classification accuracy versus topic number on three datasets. The FL setting is $\alpha=1$ and $K=10$, solid lines represent FedNMF+MI methods while dotted lines for FedNMF without MI methods. We utilize red, blue, green and purple to color four different methods: FedAvg, FedAdagrad, FedAdam and FedYogi.
Throughout this section, we apply the same plotting scheme.
In general, as the increase of topic number $K$, the text classification accuracy rises, which is due to the rapid growth of model capacity. The solid lines are consistently above the dotted lines of the same color, which shows the effectiveness of MI in the improvement of downstream classification task.

\begin{figure}[h]
\centering
\includegraphics[width=\linewidth]{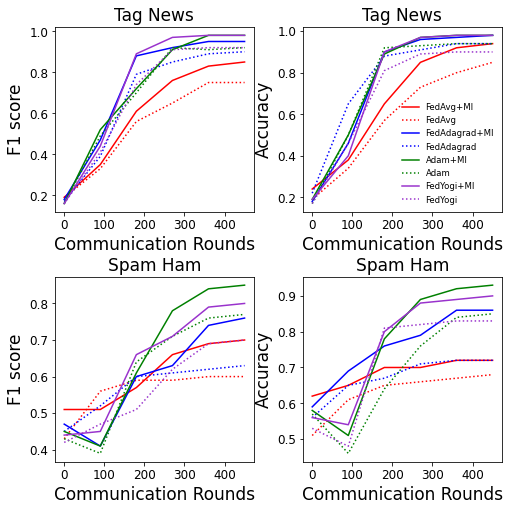}
\caption{The performance of various topic modeling on Tag.News and Spam ham versus communication cost. FL setting: $\alpha=1.0$ and $K=10$.}
\label{fig:comm}
\end{figure}
\paragraph{Communication Complexity} Communication cost is an important factor in FL, so we investigate the communication cost of our FedNMF methods under the FL setting: $\alpha=1.0$ and $K=10$. The topic number is set at 100.
Figure \ref{fig:comm} presents the F1 score of eight federated methods versus the communication cost. The solid lines represent FedNMF+MI methods while dotted lines for methods without MI. The red, blue, green and purple lines indicate FedAvg, FedAdagrad, FedAdam and FedYogi respectively.
The solid lines tends to converge faster than the dotted line of the same color, which means that MI makes the FedNMF converges faster.

\paragraph{Topic Examples Evaluation}
To qualitatively illustrate the high-quality topics
generated by federated learning approaches, Table \ref{tab:topic_words} presents the examples of topic words yielded by FedLDA, FedAvg, FedAdam, FedAvg+MI and FedAdam+MI algorithms on the Stack Overflow dataset.
This table exhibits that FedLDA tends to generate some repetitive topics with repeated words, such as ''bash script" and ''command line", and
although the topics of FedAdam and FedAvg seem diverse, they are
less informative and less coherent. The topic quality of FedAdam+MI and FedAvg+MI is apparently higher than others. This also illustrates the effectiveness of MI for FedNMF algorithms.

\begin{table}[htp]
\begin{adjustbox}{scale=0.9,tabular=ll,center}\\
\hline
Models & Topic Word Examples                            \\ \hline
       & node display specific access working category cell      \\
     & matlab matrix \underline{plot} function \underline{plotting} graph image     \\
 FedLDA  & \underline{command line bash script} shell upload folder  \\  
 & \underline{bash script command shell} \underline{scripting} \underline{run} \underline{running} \\
 & error list could give infinite syntax throwing 
 \\\hline
       & java class scala actor wordpress matlab function      \\
     & matlab matrix \underline{plot} function \underline{plotting} graph image       \\
 FedAvg  & qt application widget creator signal window gui \\  
 & bash \underline{script} command shell \underline{scripting} \underline{run} \underline{running} \\
 & bash file script excel \underline{reading} folder \underline{read} open \\
 \hline
       & wordpress matlab scala actor function java class     \\
     & matrix plot matlab function vector graph color       \\
 FedAvg  & qt gui application widget creator thread event linux\\
 +MI & svn repository branch revision external server  \\
 &  upload folder bash file script excel read open \\
       \hline
       & scala actor wordpress matlab function java class     \\
     & matlab matrix \underline{plot} function \underline{plotting} image color       \\
 FedAdam  & qt application widget creator signal window gui \\  
 & found error list warning message cause internal loop \\
 & bash file script excel \underline{reading} folder \underline{read} open\\\hline
       & visual studio solution build project debug test     \\
FedAdam       & mac type error system osx development application       \\
+MI & qt application widget creator signal window gui \\  
 & error list page message found cause internal  \\
 & file svn bash excel reading fold upload open\\
       \hline       
\end{adjustbox}
\caption{\label{tab:topic_words}Topic words examples under $K=200$. Repetitive words are underlined.}
\end{table}

\begin{figure}[h]
\centering
\includegraphics[width=0.98\linewidth]{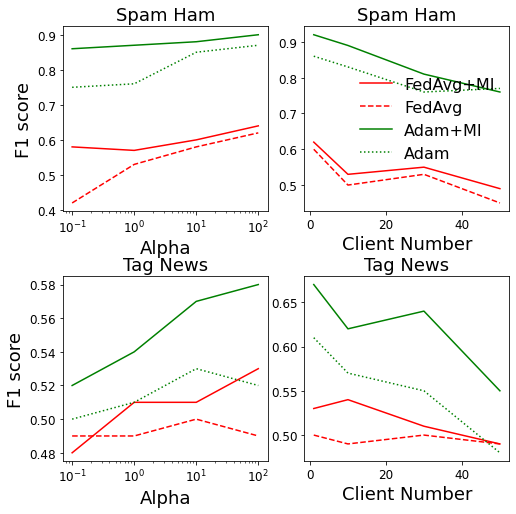}
\caption{The performance of FedAvg, FedAvg+MI, FedAdam, FedAdam+MI under different settings of $\alpha$ and $K$.}
\label{fig:alpha_client}
\vspace{-5mm}
\end{figure}

\paragraph{Varying $\alpha$ and Client Number}
We also study the effect of data heterogeneity and client number on FedNMF+MI algorithms. Figure \ref{fig:alpha_client}
displays the performance of FedAvg+MI and FedAdam+MI over varying values of $\alpha$ and client number $K$. Across two datasets, the general trend is that FedNMF performs better as $\alpha$ increases and performs worse as client number $K$ increases. This is because the data heterogeneity across clients decreases as $\alpha$ increases, so it makes it easier for federated methods (both FedNMF and FedNMF+MI) to yield well-trained topic models. As the client number $K$ increases, the data is more scattered, so it becomes harder for FedNMF(+MI) algorithms to extract good topics.

\section{Conclusion}

In this paper, we propose a framework, FedNMF, for NMF based federated topic modeling methods, which can yield high-quality topics when the documents are locally stored. To mitigate the performance degradation caused by data heterogeneity across clients, we propose the FedNMF+MI framework, which further maximizes the mutual information between the input text count features and topic weights. This sheds light on the potential use of mutual information in NMF based topic modeling. As the increasing awareness on data privacy, our FedNMF+MI algorithms can be helpful in many applications such as the decentralized short texts analysis and short document content mining.


\section*{Acknowledgment}
This paper is supported by the Key Research and Development Program of Guangdong Province under grant No.2021B0101400003. Corresponding author is Jianzong Wang from Ping An Technology (Shenzhen) Co., Ltd (jzwang@188.com).

\bibliographystyle{IEEEtran}
\bibliography{fed_nmf}

\end{document}